\documentclass[USenglish]{article}
\usepackage[utf8]{inputenc}
\usepackage{newunicodechar}
\newunicodechar{−}{--} 
\usepackage[small]{dgruyter}
\usepackage{microtype}
\usepackage{amsfonts}
\usepackage{euler}
\usepackage{amsmath}
\usepackage{booktabs}
\usepackage{colortbl}
\usepackage{tikz}
\usetikzlibrary{calc,backgrounds,external}
\usepackage{pgfplotstable}
\usepackage{pgfplots}
\usepackage[group-minimum-digits=4,
            mode=text,
            table-format=8.0,
            table-align-exponent=false,
            table-align-uncertainty=false,
            round-mode=places,
            round-precision=2]{siunitx}
\usepackage{hyperref}
\hypersetup{
colorlinks=true,
linkcolor=[rgb]{0 0 0.2},
urlcolor=[rgb]{0 0 0.2},
filecolor=[rgb]{0 0 0.2},
citecolor=[rgb]{0 0 0.2},
pdftitle={Measuring morphological complexity}
}

 \usepackage[natbib,citetracker=true,citestyle=authoryear,
             maxcitenames=2,maxbibnames=99,minnames=1,bibstyle=unified,
             backend=biber,uniquename=false,uniquelist=false,url=false]{biblatex}
\usepackage{caption}
\begin{document}

  \title{What do complexity measures measure?}
  \subtitle{Correlating and validating corpus-based measures of morphological complexity}
  \runningtitle{Measures of morphological complexity}
  \articletype{...}
  \author*[1]{Çağrı Çöltekin}
  \author[2]{Taraka Rama}
  \runningauthor{Çöltekin and Rama}
  \affil[1]{University of Tübingen,
            Department of Linguistics, Tübingen, Germany,
            E-mail: ccoltekin@sfs.uni-tuebingen.de
  }
  \affil[2]{University of North Texas, Department of Linguistics, Texas, USA,
            E-mail: taraka.kasi@gmail.com
  }
  \communicated{...}
  \dedication{...}
  \received{...}
  \accepted{...}
  \journalname{...}
  \journalyear{...}
  \journalvolume{..}
  \journalissue{..}
  \startpage{1}
  \aop
  \DOI{...}

  \abstract{We present an analysis of eight measures
    used for quantifying morphological complexity of natural languages. 
    The measures we study are corpus-based measures of morphological
    complexity with varying requirements for corpus annotation. 
    We present similarities and differences between these measures
    visually and through correlation analyses, as well as 
    their relation to the relevant typological variables.
    Our analysis focuses on whether these `measures' are
    measures of the same underlying variable,
    or whether they measure more than one dimension of
    morphological complexity.
    The principal component analysis indicates that the first principal
    component explains \SI{92.6213}{\percent} of the variation
    in eight measures, indicating a strong linear dependence
    between the complexity measures studied.
  }

\maketitle

\section{Introduction}\label{sec:intro}

Whether a language is more complex than another is an intriguing question.
It has been widely assumed that all human languages have more-or-less
equal complexity.%
\footnote{\citet[p.180]{hockett1958} is the common reference
where this hypothesis is stated clearly.
See, however, for example, \citet{sampson2009} for more evidence
of broad acceptance of the equal-complexity hypothesis.}
Recent challenges \citep{mcwhorter2001,sampson2009} to
this `equal-complexity' hypothesis resulted in a large number of studies
which aim to objectively measure complexities of human languages.
In general, ranking languages of the world on a scale of complexity
is not necessarily very productive or useful by itself.
However,
such measures are useful, and used for assessing effects of
geographic, historical, social, cultural, political and cognitive variables
on linguistic differences in specific domains of linguistic structure
\citep[e.g.,][]{mcwhorter2001,kusters2003,bulte2012,bentz2016,bozic2007,miestamo2008,szmrecsanyi2012,vainio2014,mehravari2015,yoon2017,berdicevskis2018evolang,brezina2019,chen2019,clercq2019,ehret2019,slik2019,weiss2019,michel2019}.
Since these studies use complexity metrics as a reference to 
linguistic differences based on other variables,
it is crucial to have objective, precise and well-understood metrics.

Complexity of a text from a single language can
be determined relatively consistently by the speakers of the language.
This notion of a complex language is also quantified by
measures that have been developed in a long tradition
of assessment of complexity of texts
written in the same language \citep[see][for a survey]{dubay2004}.
Creating objective measures
for comparing the complexities of different languages, however,
comes with multiple difficulties,
ranging from the lack of a clear definition of complexity \citep{andrason2014}
to the fact that it is likely impossible to summarize 
the complexity of a language with a single number \citep{deutscher2009}.
Measuring the complexity of subsystems of a language,
particularly complexity of morphology, seems less controversial.
Although it is not completely free of the issues noted for measuring  
overall complexity of a language,
the intuition that the morphological complexity of Mandarin
is less than the morphological complexity of Estonian
is hardly open to debate.
Quantifying this intuition has been an active strain of research
yielding a relatively large number of measures of morphological complexity
\citep[][just to name a few]{juola1998,dahl2004,sagot2011,newmeyer2014,bentz2017,koplenig2017,stump2017,berdicevskis2018ud,cotterell2019}.

Most complexity measures suggested in the literature are
necessarily indirect, noisy, theory- or model-dependent
and can often be applied to a limited number of languages
due to lack of resources or information.
Furthermore,
morphological complexity is argued
to have multiple dimensions \citep{anderson2015}.
As a result, understanding and validating these measures are 
crucial for research drawing conclusions based on them.
Despite the large number of seemingly different 
measures proposed for quantifying linguistic complexity
over the last few decades,
there has been only a few attempts to compare and understand these measures
\citep[e.g.,][]{bentz2016,stump2017,berdicevskis2018ud}.
In this paper,
we experiment with a number of measures of morphological complexity,
investigating their similarities and differences
as well as their relation to typological features
obtained from grammar descriptions.
In particular, given a set of measures of morphological complexity
(described in Section~\ref{sec:measures}),
we focus on the question of whether these `measures' address
the same underlying concept and construct,
and if they differ in (typographically) meaningful ways.


\section{Measuring morphological complexity}

Unlike other areas of inquiry in linguistics,
morphology is a very popular domain of language
for studying linguistic complexity.
Measuring morphological complexity is often
motivated based on the fact that
it is relatively straightforward and theory-independent,
e.g., in comparison to syntax \citep{juola1998}.
Another motivating factor is the claim that the mere
existence of morphology is complexity \citep{carstairs2010,anderson2015},
which is parallel to the claim that younger languages tend to 
have simpler morphologies \citep{mcwhorter2001}.

It is, however, often unclear what most studies mean by the complexity of
(subsystems of) languages \citep{sagot2013}.
Most typological studies quantify the morphological complexity
based on counting a set of properties.
Besides the strong consensus that different subsystems of languages
may have different complexities,
the concept of morphological complexity alone is probably a multi-faceted
concept \citep{anderson2015} which may be impossible or difficult
to place on a single scale.
A common distinction popularized by \citet{ackerman2013} is between
\emph{enumerative} and \emph{integrative} complexity.
The enumerative complexity is based on the number of 
morphosyntactic distinctions marked on words of a language,
while the integrative complexity is about the predictability
of morphologically related words from each other.
The former notion is similar to the notion of complexity
in most typological studies,
and it is inline with what computational linguists typically
call `morphologically rich' \citep[e.g.,][]{tsarfaty2013}.
The latter, however, with simplification,
indicates what one would associate with `difficulty' in processing and learning.
A language may exhibit high complexity in one of these scales,
while being less complex on the other.
For example, an agglutinating language with many possible morphosyntactic
alternations with a regular mapping between the functions and the forms
may have a high enumerative complexity but low integrative complexity.

One of the aims of the present study is to provide evidence
for this distinction.
Given a large number of measures suggested in earlier literature,
we perform a principal component analysis (PCA) to observe whether
there are more than one meaningful independent dimensions measured
by this seemingly diverse set of measures.
Earlier work on quantifying `integrative' complexity has been based on
the paradigm cell filling problem \citep{ackerman2009}.
Both \citet{ackerman2013} and \citet{cotterell2019} calculate a version
of this complexity using paradigms extracted from grammars
and lexical databases, respectively.
These studies are particularly interesting as they postulate
multiple dimensions of morphological complexity,
and methods of quantifying these dimensions.
\citet{cotterell2019} also report an inverse correlation
between these two dimensions of morphological complexity.

Another aspect of a complexity measure is the resource needed
for measuring it.
The two studies listed above use morphological information extracted from
grammar descriptions and crowd-sourced lexical data.
Corpus-based approaches are also commonly used
for quantifying linguistic complexity
\citep[e.g.,][]{juola1998,oh2013,bentz2016}.
A straightforward method of measuring morphological complexity from
an unannotated corpus is based on a  measure of lexical diversity.
Since morphologically complex languages tend
to include a larger number of word forms,
they tend to exhibit higher lexical diversity.
Another common approach is to utilize entropy of the text.
In particular,
individual words are expected to contain more regularities (lower entropy)
in a morphologically complex language.
Linguistically annotated corpora allows
formulating more direct measures of the notions of complexity discussed above.
For example, one can approximate
the enumerative complexity
by counting the available morphological features in the corpus,
and integrative complexity by training a machine learning method
to learn a mapping between the forms and the functions of words
based on the information available in the corpus.

It is important to note, however,
that there are certain differences in comparison
to the methods that work on
data extracted from grammar descriptions or lexical resources.
The information in a corpus reflects the language use,
rather than a theory or description of the language.
This means some word forms or paradigm cells will not be observed in a corpus.
On the other hand,
obtaining a corpus is often easier than extracting data from
linguistic documentation.
Hence, a corpus-based approach is more suitable for a wider range of languages.
Furthermore, a corpus also provides frequency information which can be utilized
instead of only type-based inferences one can make from lexical data.
All measures we study in this paper are corpus-based measures.

\section{Measures}\label{sec:measures}

The present study compares \num{8} different corpus-based measures of
morphological complexity.
The annotation level required by each measure differs from none
to full morphological (inflectional) annotations typically found
in a treebank.
Most of the measures we define in this section are used in earlier
studies for quantifying morphological complexity.
A few new measures are also introduced here,
but all are related to measures from earlier literature.
In some cases,
we modify an existing measure to either resolve some methodological issues,
or adapt it to current experimental setup.
The remainder of this section describes the measures we study.
The details of the experimental setup is described
in Section~\ref{sec:experiments}.

\paragraph*{Type/token ratio (TTR)}
The \emph{type/token ratio}
is the ratio of word types (unique words) to word tokens in a given text sample.
The TTR is a time-tested metric for measuring linguistic complexity.
Although there has been criticisms of using TTR
as a measure of lexical diversity \citep{jarvis2002,mccarthy2010},
it is one of the most straightforward measures to calculate,
and it has been used  in a number of earlier studies
for measuring morphological complexity
that showed rather high correlation with other,
more complex methods \citep[e.g.,][]{bentz2016,coltekin2018mlc,cech2018}.
Since morphologically complex languages have more diverse word forms,
high TTR indicates rich or complex morphology.
Since the TTR depends on the corpus size,
it is a common practice to calculate the TTR using a fixed window size
\citep{kettunen2014}.%
\footnote{Although there are also many instances of the use of TTR without
  controlling corpus size in the literature.
}
We calculate the TTR on a fixed-length random sample,
and report the average over multiple samples.
The details of the sampling procedure is described in Section~\ref{sec:data}.

\paragraph*{Information in word structure (WS)}
A popular framework for measuring morphological complexity
is based on the information
between an original text and a distorted version of the text
where the word structure is destroyed.
The measure was initially proposed by \citet{juola1998}
but variations of the framework has been used in a large number of studies
\citep[, just to name a few]{juola2008,montemurro2011,ehret2014,bentz2016,ehret2016,koplenig2017}.
The general idea of the measure is that the difference of entropy between
the original version and the distorted version is related to
the information expressed by the morphology of the language.

In this work,
we follow \citet{juola1998} and use `compressibility' as a measure
of (the lack of) information.
In other words,
we expect worse compression ratios for a distorted text
in comparison to its non-distorted version in a morphologically complex language.
However,
instead of compressed file ratios used by some of the earlier work,
we take the difference of compression ratios between the original and 
the distorted text as the measure of complexity
(similar to \cite{koplenig2017} and \cite{bentz2016}).
Crucially, we replace every word type in the corpus
with the same random sequence with the equal length.
To preserve some of the phonological
(more precisely orthographic) information,
we do not generate the random `words' uniformly,
but from a unigram language model of letters estimated from the corpus.
Note that the measure depends on the units used for measuring entropy.
As a result, it is not meaningful to use it for comparing
texts written with different writing systems.

\paragraph*{Word and lemma entropy (WH, LH)}
The entropy of the word-frequency distribution has also been used
as a measure of morphological complexity in the earlier literature.
\citet{bentz2016} motivates the measure as
the average information content of a word in the text sample studied.
Another interpretation of the score is based on the typical distributions 
of words observed in languages with varying morphological complexity.
The word distributions are affected by two issues related to morphology.
First, complex morphology creates many rare word forms,
resulting in a longer tail of the word frequency distribution,
and hence, less predictable words.
Second, a morphologically `poor' language typically uses more function words,
resulting in more words with higher probabilities,
and hence high predictability and low entropy.
The measure of word entropy we use is similar to the one used by 
\citet{bentz2016}.
However, we use the maximum-likelihood estimates of the word probabilities
in the entropy calculation.%
\footnote{As noted in earlier studies, the maximum-likelihood estimate
 overestimates the observed word probabilities
 (hence, underestimates the entropy).
 In our experiments, smoothed probability estimates did not
 yield any recognizable differences,
 likely because (1) we are not interested in entropy but the differences
 between entropy values, and (2) controlling the size of the corpus
 yields the same amount of overestimation for each text.
}
Furthermore,
similar to the other metrics,
we calculate the word entropy on a fixed-sized sample for all languages
to remove potential effects of the sample size.

The interpretation offered above for the word entropy suggests
two separate effects.
Since we work with a data set including lemma annotations,
we also calculate the lemma entropy,
which should be less sensitive
to the information packed in the words on average,
but bring the effect due to the large number of function words to the front.
The frequencies of content lemmas are
expected to be relatively stable across languages
with different morphological complexities.
Since lemmatization strips out the inflections from the words,
the differences observed in lemma entropy across languages are
likely to be because of the frequencies of function words,
and rich derivation and compounding.
Since derivation and compounding increase the inventory of lemmas,
languages with rich derivational morphology and compounding 
are expected to get higher LH scores.
To our knowledge, the lemma entropy is not considered
in the earlier literature in this form.
However, `lexical predictability' measure of \citet{blache2011}
(the ratio of frequent lemmas to all lemmas) is related to our measure.

\paragraph*{Mean size of paradigm (MSP)}
The \emph{mean size of paradigm} is the number of word-form types divided by
the number of lemma types.
Our calculations follow \citet{xanthos2011}
who use MSP to correlate morphological complexity 
and the acquisition of morphology during first language acquisition.
If the morphology of a language has a large number of paradigm cells
(and if those complex paradigms are used in real-world language),
then MSP will be high.
The same measure is termed `morphological variety' by \citet{blache2011}.

\paragraph*{Inflectional synthesis (IS)}
The \emph{index of synthesis of the verb} \citep{comrie1989}
is a typological measure of morphological complexity,
which is also used by \citet{shosted2006}
for investigating the correlation between morphological
and phonological complexity.
\citet{shosted2006} uses the measure
as extracted from grammar descriptions by \citet{bickel2005}.
The measure is the number of inflectional features a verb can take.
Here, we adapt it to a corpus-based approach.
Our version is simply the maximum number of distinct inflectional features
assigned to a lemma in the given sample.
Some systematic differences are expected due to the linguistic coding
in UD treebanks and the way \citet{bickel2005} code morphological features.
For example, \citet{bickel2005} do not consider fused morphemes
as separate categories,
e.g., a fused tense/aspect/modality (TAM) marker counted only once,
while treebanks are likely to code each TAM dimension
as a separate morphological feature.

\paragraph*{Morphological feature entropy (MFH)}

The limitation of the inflectional synthesis measure described above to
count the morphological features only on verbs is likely due to the cost of
collecting data from descriptive grammars of a large number of languages.
Having annotated corpora for many languages
allows a more direct estimate of
the use of inflectional morphology in the language.
To also include the information on usage,
we calculate the entropy of the feature--value pairs,
similar to the WH and LH measures described above.
Everything else being equal,
the measure will be high for languages with many inflectional features.
However, this measure is affected by language use.
For example, a rarely used feature value, e.g.,
use of a rare/archaic case value,
will not affect the value of this measure
as much as a set of uniformly used case values.

\paragraph*{Inflection accuracy (IA)}
The \emph{inflection accuracy} metric we use in this study is
simply the accuracy of a machine learning model
predicting the inflected word from its lemma and morphological features.
The intuition is that if the language in question has a complex morphology,
the accuracy is expected to be low.
As a result, we report the negative inflection accuracy
(indicated as \emph{--ia}) in the results below.

Unlike the measures discussed above,
inflection accuracy is expected to be high for languages
if the language has a rather regular, transparent inflection system
-- even if it utilizes a large number of inflections.
In other words,
this measure should be similar to integrative complexity 
of \citet{ackerman2013}.
Unlike \citet{ackerman2013} and \citet{cotterell2019},
we estimate the inflection system from
word tokens rather than word types,
which results in incomplete paradigms in comparison
to models trained on lexical data.

The inflection method used in this study is based on linear classifiers,
which provides close to state-of-the-art systems
with relatively small demand on computing power.
A neural-network-based inflection system
\citep[common in recent SIGMORPHON shared tasks, e.g.,][]{sigmorphon2018st,sigmorphon2019st}
may provide higher accuracies.
However, our focus here is on the differences in inflection accuracy
rather than the overall success of the system in the inflection task.
There is no a priori reason to expect substantial differences
when scores on different languages are compared.
We use the freely available inflection implementation by 
\citet{coltekin2019sigmorphon} in this study.

\section{Data and experimental setup}\label{sec:experiments}

\subsection{Data}\label{sec:data}

Our data consists of \num{63} treebanks
from the Universal Dependencies (UD) project \citep{nivre2016}.
The set of treebanks were selected for
the Workshop on Measuring Linguistic Complexity (MLC 2019).%
\footnote{The data is based on UD treebanks version 2.3 \citep{ud2.3},
  and can be accessed at \url{http://www.christianbentz.de/MLC_data.html}}
The full list of treebanks, along with statistics are provided in
Table~\ref{tbl:data} in Appendix~\ref{sec:appendix}.
Most treebanks are from the Indo-European language family
(\num{51} of the \num{63} treebanks).
15 languages are represented by multiple treebanks in the data set
which are helpful for distinguishing differences across languages
and differences due to text types from the same language.
Some treebanks/languages do not include morphological annotations,
and the sizes of treebanks are highly variable.
The smallest treebank (Hungarian) has about \num{40}K tokens,
while the largest (Czech PDT) consists of
approximately \num[round-precision=1]{1.5}M tokens.

The usage of UD POS tag inventory is relatively stable across languages.
The number of POS tags used vary between \num{14} and \num{18}.
The morphological features in different treebanks are more varied,
ranging between \num{2} to \num{29} feature labels.

Since some of our measures require morphological features,
we exclude the treebanks without morphological features
(notably Japanese and Korean treebanks) from most of our analyses.%
\footnote{These treebanks include some sporadic morphological features marked,
e.g., the Japanese treebank marks cardinal numbers.
However, no linguistically interesting features were marked
despite the fact that both languages are morphologically complex.}
Furthermore, since some of the other features depend on the writing system,
we also exclude treebanks of languages with non-alphabetic writing systems,
which also excludes Chinese treebank in the collection.

For testing the relevance of the measures with typological variables,
we use a set of \num{28} typological variables related to morphology
from the World Atlas of Language Structures
\citep[WALS,][]{wals}.
The same set of features are also used by \citet{bentz2016}.
Note that not all features are available for all languages in our sample.
The list of features and their coverage is listed in
Table~\ref{tbl:wals} in Appendix~\ref{sec:appendix}.

\subsection{Experimental setup}

As noted above, some of our measures depend on text size.
For comparability,
we calculate all measures on \num{20000} tokens
(approximately half of the smallest treebank)
sampled randomly from the input treebank.
Since some of the measures (e.g., WS) are sensitive to word order,
our sampling process samples sentences randomly with replacement
until the number of tokens reach \num{20000}.
For all measures except inflection accuracy,
we repeat this process \num{100} times,
and report the mean scores obtained over these random samples.%
\footnote{The source code used for calculating the measures is publicly available at \url{https://github.com/coltekin/mcomplexity}.}

The inflection accuracy is measured on the same data set,
the model is trained and tested on the inflection tables extracted from
a single \num{20000}-token sample randomly obtained from each treebank.
However, due to computational reasons,
we do not repeat the process multiple times but 
report the average score over cross validation experiments.
Specifically,
the score we present are the best mean accuracy
(exact match of the inflected word)
obtained through 3-fold cross validation on this sample.
The model is tuned for each language separately
using a random search through the model parameters.

Since there is no gold-standard for evaluating a complexity metric,
we present all values graphically,
which serves as an informal validation
(a measure that puts Finnish and Russian high on the scale,
while assigning lower complexities to English and Vietnamese
is probably measuring something relevant to morphological complexity).
The graphs also allow a visual inspection of differences
between languages and between measures.
We also present linear and rank-based correlation coefficients
between the measures to quantify the relations between the individual measures.

We further validate the measures by evaluating their relationship
with the morphology-related features from WALS.
Since WALS features are categorical,
instead of assigning an ad hoc numeric score to a configuration of features,
we use ridge regression
\citep[as implemented in \texttt{scikit-learn};][]{pedregosa2011scikit}
to predict normalized complexity measures from selected WALS features.
If the complexity measure is relevant to one or more of the variables,
the prediction error by the regression model will be low. 
To prevent overfitting,
we tune and test the regression model using leave-one-out cross validation.
For each measure,
we present the reduction of average error (in comparison to a random baseline
whose expected root-mean-squared error is \num{1} on a standardized variable)
as a measure of relation to the WALS features.

To analyze whether measures include multiple dimensions or not,
we perform dimensionality reduction using principal component analysis (PCA).
The intuition here is that if the measures differ in what they measure,
the explained variance should be shared among multiple principal components.
Furthermore, if  the lower-order principal components measure
meaningful dimensions of morphological complexity,
we expect them to indicate linguistically relevant differences
between languages.

\section{\label{sec:results}Results}

Figure~\ref{fig:measures} presents the values of each complexity measure
on the x-axis of the corresponding panel and the language/treebank codes displayed
in each panel are sorted by the corresponding complexity score.
Note that the points in these graphs represent averages over 
\num{100} bootstrap samples.
The standard deviations of the scores obtained on multiple samples
are rather small and are not visible on the graphs when plotted.

In general,
all measures seem to show  the expected trend:
morphologically complex languages are generally on the top of all scales
(mainly agglutinative languages like Finnish, Turkish,
but also fusional ones like Russian and Latin).
Similarly, languages like Vietnamese, English and Dutch are generally
at the lower end of the scale.
Figure~\ref{fig:measures} also includes scores
for the treebanks with missing morphological annotations
(Japanese and Korean treebanks).
For the measures that do not require morphological annotations,
these languages are placed close to the top of the scale.
However, when morphological annotations are required,
they are ranked at the bottom.
Since the WS score is sensitive to the writing system,
the ranking is meaningful only for languages with sufficiently similar
writing systems.

\begin{figure}
  \tikzsetnextfilename{fig-measures}%
  \input{all-measures-fig.tex}
  \vspace{-5mm}
  \caption{A visualization of all measures on all treebanks.
    Languages are sorted according to the measure value in each panel
    (higher points indicate higher complexity).
    The original scale of each measure is given under each panel.
    The vertical gray lines represent the mean value of the measure.
  }\label{fig:measures}
\end{figure}

Table~\ref{tbl:corr} reports the correlations between the measures.
The correlations are almost always positive, and quite strong for many pairs.
In contrast to the findings of \citet{cotterell2019},
we do not observe any negative correlation between
any of the measures and the negative inflection accuracy
(which we expect to measure `integrative complexity' to some extent).
A potential reason for this is the fact that our model is trained on
inflection tables extracted from a corpus.
Our inflection tables are sparse where
infrequent paradigm cells are not present and
paradigms of infrequent words are underrepresented in the input of our model.
As a result, our model/measure is affected more by frequent forms, 
which are also expected to include more irregular forms.

\begin{table}
\centering
  \caption{Correlations between the measures.
    The lower triangular matrix reports
    linear (Pearson) correlation coefficients.
    The upper triangular matrix reports
    rank (Spearman) correlation coefficients.
    Darker shades indicate stronger correlation.
    All correlations, except the ones marked with an asterisk,
    are significant at $p < 0.05$.
  }\label{tbl:corr}
  \tikzexternaldisable
  \setlength{\tabcolsep}{1pt}
\renewcommand{\arraystretch}{0}
\begin{tabular}{l*{8}{r}}
& {ttr} & {msp} & {ws} & {wh} & {lh} & {is} & {mfh} & {-ia}  \\
\midrule
ttr
& \phantom{|}& \tikz{\node[align=right,text width=9mm,minimum height=4ex,anchor=base,fill=blue!30.75994340028916] {$0.41^{\text{\tiny\phantom{*}}}$};}& \tikz{\node[align=right,text width=9mm,minimum height=4ex,anchor=base,fill=blue!22.357654803285246] {$0.30^{\text{\tiny\phantom{*}}}$};}& \tikz{\node[align=right,text width=9mm,minimum height=4ex,anchor=base,fill=blue!63.84308960595528] {$0.85^{\text{\tiny\phantom{*}}}$};}& \tikz{\node[align=right,text width=9mm,minimum height=4ex,anchor=base,fill=blue!57.895505859915716] {$0.77^{\text{\tiny\phantom{*}}}$};}& \tikz{\node[align=right,text width=9mm,minimum height=4ex,anchor=base,fill=blue!37.70240813749341] {$0.50^{\text{\tiny\phantom{*}}}$};}& \tikz{\node[align=right,text width=9mm,minimum height=4ex,anchor=base,fill=blue!30.43695591989911] {$0.41^{\text{\tiny\phantom{*}}}$};}& \tikz{\node[align=right,text width=9mm,minimum height=4ex,anchor=base,fill=blue!45.13750038450891] {$0.60^{\text{\tiny\phantom{*}}}$};}\\
msp
& \tikz{\node[align=right,text width=9mm,minimum height=4ex,anchor=base,fill=blue!16.97361176338631] {$0.23^{\text{\tiny*}}$};}& \phantom{|}& \tikz{\node[align=right,text width=9mm,minimum height=4ex,anchor=base,fill=blue!30.192408256175213] {$0.40^{\text{\tiny\phantom{*}}}$};}& \tikz{\node[align=right,text width=9mm,minimum height=4ex,anchor=base,fill=blue!37.948721892399035] {$0.51^{\text{\tiny\phantom{*}}}$};}& \tikz{\node[align=right,text width=9mm,minimum height=4ex,anchor=base,fill=blue!6.508197729859424] {$0.09^{\text{\tiny*}}$};}& \tikz{\node[align=right,text width=9mm,minimum height=4ex,anchor=base,fill=blue!47.58235355461037] {$0.63^{\text{\tiny\phantom{*}}}$};}& \tikz{\node[align=right,text width=9mm,minimum height=4ex,anchor=base,fill=blue!32.38872312282753] {$0.43^{\text{\tiny\phantom{*}}}$};}& \tikz{\node[align=right,text width=9mm,minimum height=4ex,anchor=base,fill=blue!68.41566950690579] {$0.91^{\text{\tiny\phantom{*}}}$};}\\
ws
& \tikz{\node[align=right,text width=9mm,minimum height=4ex,anchor=base,fill=blue!25.71424092771363] {$0.34^{\text{\tiny\phantom{*}}}$};}& \tikz{\node[align=right,text width=9mm,minimum height=4ex,anchor=base,fill=blue!23.57346449583283] {$0.31^{\text{\tiny\phantom{*}}}$};}& \phantom{|}& \tikz{\node[align=right,text width=9mm,minimum height=4ex,anchor=base,fill=blue!21.513273247408414] {$0.29^{\text{\tiny\phantom{*}}}$};}& \tikz{\node[align=right,text width=9mm,minimum height=4ex,anchor=base,fill=blue!12.907963948445046] {$0.17^{\text{\tiny*}}$};}& \tikz{\node[align=right,text width=9mm,minimum height=4ex,anchor=base,fill=blue!17.507827451067097] {$0.23^{\text{\tiny*}}$};}& \tikz{\node[align=right,text width=9mm,minimum height=4ex,anchor=base,fill=blue!10.250238395521242] {$0.14^{\text{\tiny*}}$};}& \tikz{\node[align=right,text width=9mm,minimum height=4ex,anchor=base,fill=blue!26.413454735611683] {$0.35^{\text{\tiny\phantom{*}}}$};}\\
wh
& \tikz{\node[align=right,text width=9mm,minimum height=4ex,anchor=base,fill=blue!68.71033583421799] {$0.92^{\text{\tiny\phantom{*}}}$};}& \tikz{\node[align=right,text width=9mm,minimum height=4ex,anchor=base,fill=blue!26.106784553109982] {$0.35^{\text{\tiny\phantom{*}}}$};}& \tikz{\node[align=right,text width=9mm,minimum height=4ex,anchor=base,fill=blue!25.402334828816354] {$0.34^{\text{\tiny\phantom{*}}}$};}& \phantom{|}& \tikz{\node[align=right,text width=9mm,minimum height=4ex,anchor=base,fill=blue!65.46264111476822] {$0.87^{\text{\tiny\phantom{*}}}$};}& \tikz{\node[align=right,text width=9mm,minimum height=4ex,anchor=base,fill=blue!44.78230349607337] {$0.60^{\text{\tiny\phantom{*}}}$};}& \tikz{\node[align=right,text width=9mm,minimum height=4ex,anchor=base,fill=blue!29.444922944415396] {$0.39^{\text{\tiny\phantom{*}}}$};}& \tikz{\node[align=right,text width=9mm,minimum height=4ex,anchor=base,fill=blue!47.02928419822203] {$0.63^{\text{\tiny\phantom{*}}}$};}\\
lh
& \tikz{\node[align=right,text width=9mm,minimum height=4ex,anchor=base,fill=blue!59.6543691295562] {$0.80^{\text{\tiny\phantom{*}}}$};}& \tikz{\node[align=right,text width=9mm,minimum height=4ex,anchor=base,fill=red!10.326521664785078] {$-0.14^{\text{\tiny*}}$};}& \tikz{\node[align=right,text width=9mm,minimum height=4ex,anchor=base,fill=blue!14.048155994792257] {$0.19^{\text{\tiny*}}$};}& \tikz{\node[align=right,text width=9mm,minimum height=4ex,anchor=base,fill=blue!64.2995826817788] {$0.86^{\text{\tiny\phantom{*}}}$};}& \phantom{|}& \tikz{\node[align=right,text width=9mm,minimum height=4ex,anchor=base,fill=blue!26.19237577465505] {$0.35^{\text{\tiny\phantom{*}}}$};}& \tikz{\node[align=right,text width=9mm,minimum height=4ex,anchor=base,fill=blue!18.88784644252361] {$0.25^{\text{\tiny*}}$};}& \tikz{\node[align=right,text width=9mm,minimum height=4ex,anchor=base,fill=blue!19.473838014088408] {$0.26^{\text{\tiny\phantom{*}}}$};}\\
is
& \tikz{\node[align=right,text width=9mm,minimum height=4ex,anchor=base,fill=blue!44.34159408328114] {$0.59^{\text{\tiny\phantom{*}}}$};}& \tikz{\node[align=right,text width=9mm,minimum height=4ex,anchor=base,fill=blue!38.591153695277846] {$0.51^{\text{\tiny\phantom{*}}}$};}& \tikz{\node[align=right,text width=9mm,minimum height=4ex,anchor=base,fill=blue!15.54610421999293] {$0.21^{\text{\tiny*}}$};}& \tikz{\node[align=right,text width=9mm,minimum height=4ex,anchor=base,fill=blue!49.01012348049561] {$0.65^{\text{\tiny\phantom{*}}}$};}& \tikz{\node[align=right,text width=9mm,minimum height=4ex,anchor=base,fill=blue!26.05693043397355] {$0.35^{\text{\tiny\phantom{*}}}$};}& \phantom{|}& \tikz{\node[align=right,text width=9mm,minimum height=4ex,anchor=base,fill=blue!27.78084347920908] {$0.37^{\text{\tiny\phantom{*}}}$};}& \tikz{\node[align=right,text width=9mm,minimum height=4ex,anchor=base,fill=blue!44.6574457725713] {$0.60^{\text{\tiny\phantom{*}}}$};}\\
mfh
& \tikz{\node[align=right,text width=9mm,minimum height=4ex,anchor=base,fill=blue!28.983917224754048] {$0.39^{\text{\tiny\phantom{*}}}$};}& \tikz{\node[align=right,text width=9mm,minimum height=4ex,anchor=base,fill=blue!26.732055654255575] {$0.36^{\text{\tiny\phantom{*}}}$};}& \tikz{\node[align=right,text width=9mm,minimum height=4ex,anchor=base,fill=blue!19.492728075016792] {$0.26^{\text{\tiny\phantom{*}}}$};}& \tikz{\node[align=right,text width=9mm,minimum height=4ex,anchor=base,fill=blue!18.790381623747383] {$0.25^{\text{\tiny*}}$};}& \tikz{\node[align=right,text width=9mm,minimum height=4ex,anchor=base,fill=blue!2.911143829284511] {$0.04^{\text{\tiny*}}$};}& \tikz{\node[align=right,text width=9mm,minimum height=4ex,anchor=base,fill=blue!31.184782407104308] {$0.42^{\text{\tiny\phantom{*}}}$};}& \phantom{|}& \tikz{\node[align=right,text width=9mm,minimum height=4ex,anchor=base,fill=blue!38.677750776707995] {$0.52^{\text{\tiny\phantom{*}}}$};}\\
-ia
& \tikz{\node[align=right,text width=9mm,minimum height=4ex,anchor=base,fill=blue!47.24087665998663] {$0.63^{\text{\tiny\phantom{*}}}$};}& \tikz{\node[align=right,text width=9mm,minimum height=4ex,anchor=base,fill=blue!58.73777255952213] {$0.78^{\text{\tiny\phantom{*}}}$};}& \tikz{\node[align=right,text width=9mm,minimum height=4ex,anchor=base,fill=blue!29.75788737640737] {$0.40^{\text{\tiny\phantom{*}}}$};}& \tikz{\node[align=right,text width=9mm,minimum height=4ex,anchor=base,fill=blue!45.45092581611679] {$0.61^{\text{\tiny\phantom{*}}}$};}& \tikz{\node[align=right,text width=9mm,minimum height=4ex,anchor=base,fill=blue!15.836280138336464] {$0.21^{\text{\tiny*}}$};}& \tikz{\node[align=right,text width=9mm,minimum height=4ex,anchor=base,fill=blue!45.07517843365973] {$0.60^{\text{\tiny\phantom{*}}}$};}& \tikz{\node[align=right,text width=9mm,minimum height=4ex,anchor=base,fill=blue!45.22525528640209] {$0.60^{\text{\tiny\phantom{*}}}$};}& \phantom{|}\\
\end{tabular}
\setlength{\tabcolsep}{6pt}
\renewcommand{\arraystretch}{1}

  \tikzexternalenable
\end{table}

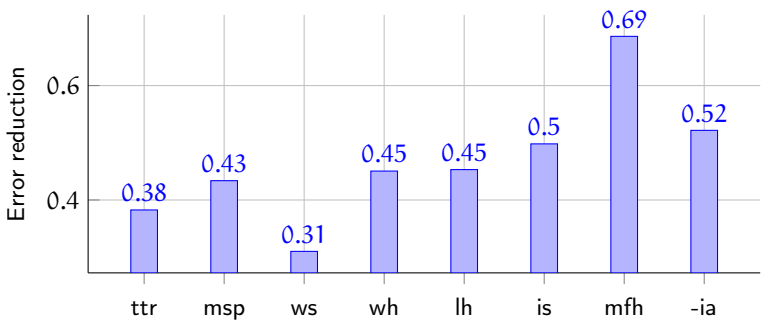
\begin{figure}%
  \centering%
  \pgfplotstableread{error.txt}\etable%
  \tikzset{external/export next=false}%
  \tikzsetnextfilename{fig-wals}%
  \begin{tikzpicture} 
    \begin{axis}[ 
        width=0.9\linewidth,
        height=5cm,
        ybar,
        axis lines*=left,
        xtick=data,
        xticklabels from table={\etable}{measure},
        xticklabel style={anchor=base, yshift=-3ex,font=\sf},
        ylabel={Error reduction},
        ylabel style={yshift=-2ex,font=\sf},
        grid,
        nodes near coords,
        /pgf/number format/fixed,
        /pgf/number format/precision=2,
      ] 
      \addplot table[x expr=\coordindex, y expr=1-\thisrow{error}]{\etable};
    \end{axis}
  \end{tikzpicture}
  \caption{The average reduction of prediction error
    when predicting the complexity measures from the WALS features.
    Higher values indicate stronger affinity with WALS features.
  }\label{fig:wals}
\end{figure}

To give an indication of the typological relevance of each measure,
we present the reduction of error in predicting each complexity 
measure from WALS morphological variables (listed in table \ref{tbl:wals}) in Figure~\ref{fig:wals}.
In general,
all complexity measures seem to be related to the typological features. 
However, since not all morphology-related features in WALS
indicate complexity,
this is also an approximate indication of validity of the measures.
Furthermore, the WALS data has many missing features for the 
languages in our data set,
reducing the value of the comparison even further.
However,
the consistently positive effect of features is a clear indication that
there is a relation between the typological features,
and the measures evaluated in this study.

\begin{figure}%
  \tikzsetnextfilename{fig-pca}%
  \centering%
  \input{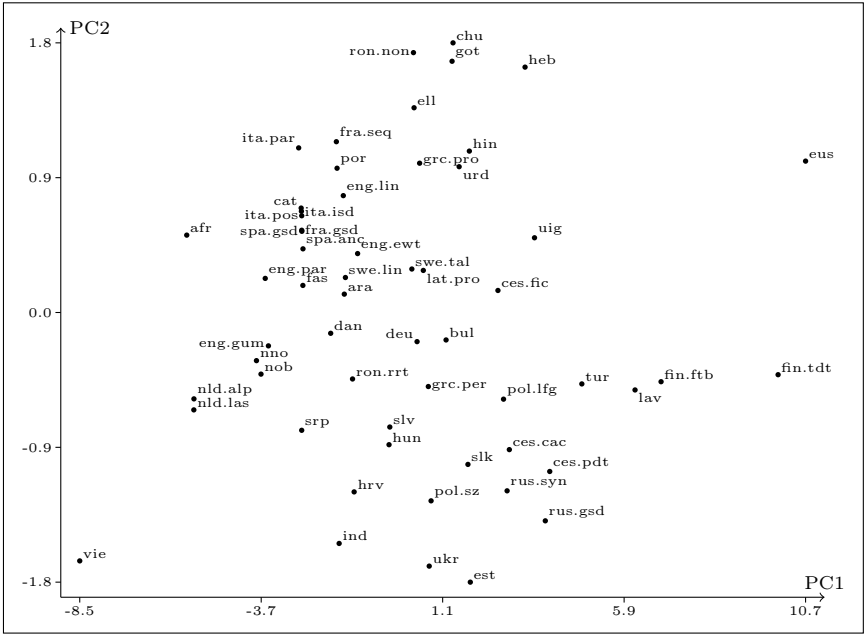}
  \caption{The first two dimensions after PCA transformation.
    The first component (x-axis) explains
    \SI{92.6213}{\percent} of the variation in the data,
    while the first two components together
    explain \SI{97.88801764344056}{\percent} of the variation.
    Note that scales differ: the y-axis is stretched for clarity.
  }\label{fig:pca}
\end{figure}

Despite overall positive correlations,
in some cases the correlations are low.
Furthermore, although correlated measures are reassuring,
if we are actually measuring different aspects of morphological complexity,
meaningful differences between the measures are more interesting.
First, to understand whether we are measuring multiple
underlying constructs, we perform a PCA analysis.
The PCA indicates that \SI{92.6213}{\percent} of the variation
in the results are explained by a single component,
suggesting a single underlying dimension.
To test this further,
we plot the first two principal components in Figure~\ref{fig:pca}.
The two dimensions together explain \SI{97.88801764344056}{\percent}
of the variation in the data.

The plot in Figure~\ref{fig:pca} clearly shows that the first PCA component
is in agreement with the intuitive notion of morphological complexity.
The languages are ranked according to expectations without many exceptions.
The languages like Basque, Finnish, Latvian and Turkish,
from different language families
but all known for their morphological complexity,
are on the higher part of the scale of the first principal component.
On the very low end of the scale, not surprisingly,
is Vietnamese, followed by Dutch and Afrikaans.
The middle part of the scale is difficult to interpret.
However we still see close clusters of languages that belong
to the same language family
and/or languages with similar morphological complexity.
The figure also shows that
multiple treebanks from the same languages are also
placed very close to each other.
Furthermore,
the relation of the first principal component to WALS features is also strong
(\num{.2836609842680956} according to the scale in Figure~\ref{fig:wals}).
However,
the second component does not provide any discernible  
indication of morphological complexity.
The treebanks from the same or similar languages are often
far apart in the second dimension,
and we do not observe any clear patterns that can be 
generalized from the second principal component.
Looking further into the principal components
does not reveal any pattern in lower-ranked components either
(the ranking of languages according to all PCA dimensions
are provided in Figure~\ref{fig:pca-measures}).
In summary,
the measures in this study seem to indicate a single underlying variable.

\section{Concluding remarks}

We presented an analysis of 8 corpus-based measures of morphological complexity.
As shown in Figure~\ref{fig:measures},
noise and minor exceptions aside,
all measures capture a sense of morphological complexity:
languages known to be morphologically complex are placed high up in all scales,
while languages that are morphologically less complex are ranked low.
Furthermore,
the treebanks that belong to the same or closely related languages obtain
similar scores.
There are some interesting differences that can also be observed here.
For example, since the measures WH and LH are sensitive to derivational
morphology and compounding,
languages like Vietnamese and English which otherwise get lower scores
obtain moderately higher scores on these metrics.

We also presented correlations between the scores (Table~\ref{tbl:corr}).
The scores generally correlate positively. 
This is reassuring, since they all intend to measure the same construct.
However, an interesting finding here is the positive correlation 
between all measures and the negative inflection accuracy.
Since inflectional accuracy is a measure of `difficulty',
based on earlier findings in the literature \citep{cotterell2019}
we expect a negative correlation between enumerative complexity measures.
We note, however,
that the data sets on which the systems are trained are different.
The inflection tables we use are extracted from corpora.
They are incomplete, and reflect the language use,
in contrast to inflection tables that include theoretically possible,
but rarely attested word forms.
As a result,
our model is trained and tested on more frequent forms
which are also more likely to be formed
by irregular, unpredictable morphological processes.
These frequent, irregular forms are likely to be overwhelmed
by many regular forms
in the complete inflection table of a language with high enumerative complexity.
In a table extracted from a relatively small corpus of the language,
the irregular forms are expected to have a larger presence.
Hence, when the model is tested on frequent words,
the difficult forms are similarly distributed regardless
of the enumerative complexity.
When tested on full inflection tables,
the effect of frequent, irregular forms diminishes for languages
with high enumerative complexity.
In other words,
perhaps languages get similar budgets for morphological irregularities
(hence integrative complexity),
but the size of the inflection tables determine the impact of these
irregularities on average predictability of forms of the inflected words
from morphological features.
Even though this explanation needs further investigation to be confirmed,
it is also supported by the fact that MSP is the measure with the highest
correlation with the negative inflection accuracy.

Finally,
despite the expectation of multiple dimensions of morphological complexity,
dimensionality reduction experiments indicate that the measures
analyzed here are likely to measure a single underlying dimension.
This is not to say that morphological complexity is uni-dimensional.
The findings indicate that if the measures at hand are measuring different
linguistic dimensions,
these dimensions are highly (positively or negatively) correlated with others.
A practical side effect of the dimensionality reduction is, however,
that the resulting single dimension seems to reflect the intuitions about
morphological complexities of the measures better,
placing the same or related languages much closer to each other.
This also suggests that when available,
combining multiple measures provides a more stable and reliable indication 
of morphological complexity compared to a single measure.

\printbibliography

\appendix
\clearpage
\section{Data and additional visualizations}\label{sec:appendix}

\noindent%
\captionof{table}{The information on treebanks used in this study.
  Size is given in \num{1000} tokens. \label{tbl:data}
}
\centering
\resizebox{!}{8cm}{
\begin{filecontents*}{treebanks.txt}
id	Language	Treebank	Size	Family	Notes
afr	Afrikaans	AfriBooms	49	Indo-European	
ara	Arabic	PADT	282	Afroasiatic	
bul	Bulgarian	BTB	156	Indo-European	
cat	Catalan	AnCora	531	Indo-European	
ces_cac	Czech	CAC	494	Indo-European	
ces_fic	Czech	FicTree	167	Indo-European	
ces_pdt	Czech	PDT	1506	Indo-European	
chu	Old Church Slavonic	PROIEL	57	Indo-European	ancient
cmn	Chinese	GSD	123	Sino-Tibetan	few features
dan	Danish	DDT	100	Indo-European	
deu	German	GSD	292	Indo-European	
ell	Greek	GDT	63	Indo-European	
est	Estonian	EDT	434	Uralic	
eng_ewt	English	EWT	254	Indo-European	
eng_gum	English	GUM	80	Indo-European	
eng_lin	English	LinES	82	Indo-European	
eng_par	English	ParTUT	49	Indo-European	
eus	Basque	BDT	121	Basque	
fas	Persian	Seraji	152	Indo-European	
fin_ftb	Finnish	FTB	159	Uralic	
fin_tdt	Finnish	TDT	202	Indo-European	
fra_gsd	French	GSD	400	Indo-European	
fra_seq	French	Sequoia	70	Indo-European	
got	Gothic	PROIEL	55	Indo-European	ancient
grc_per	Ancient Greek	Perseus	202	Indo-European	ancient
grc_pro	Ancient Greek	PROIEL	214	Indo-European	ancient
heb	Hebrew	HTB	161	Afroasiatic	
hin	Hindi	HDTB	351	Indo-European	
hrv	Croatian	SET	197	Indo-European	
hun	Hungarian	Szeged	42	Indo-European	
ind	Indonesian	GSD	121	Austronesian	
ita_isd	Italian	ISDT	298	Indo-European	
ita_par	Italian	ParTUT	55	Indo-European	
ita_pos	Italian	PoSTWITA	124	Indo-European	
jpn	Japanese	GSD	184	Japonic	few features
kor_gsd	Korean	GSD	80	Indo-European	few features
kor_kai	Korean	Kaist	350	Koreanic	no features
lat_itt	Latin	ITTB	353	Indo-European	ancient
lat_pro	Latin	PROIEL	199	Indo-European	ancient
lav	Latvian	LVTB	152	Indo-European	
nld_alp	Dutch	Alpino	228	Indo-European	
nld_las	Dutch	LassySmall	98	Indo-European	
nno	Norwegian	Nynorsk	301	Indo-European	
nob	Norwegian	Bokmaal	310	Indo-European	
pol_lfg	Polish	LFG	130	Indo-European	
pol_sz	Polish	SZ	83	Indo-European	
por	Portuguese	Bosque	227	Indo-European	
ron_non	Romanian	Nonstandard	195	Indo-European	
ron_rrt	Romanian	RRT	218	Indo-European	
rus_gsd	Russian	GSD	99	Indo-European	
rus_syn	Russian	SynTagRus	1107	Indo-European	
slk	Slovak	SNK	106	Indo-European	
slv	Slovenian	SSJ	140	Indo-European	
spa_anc	Spanish	AnCora	549	Indo-European	
spa_gsd	Spanish	GSD	431	Indo-European	
srp	Serbian	SET	86	Indo-European	
swe_lin	Swedish	LinES	79	Indo-European	
swe_tal	Swedish	Talbanken	96	Indo-European	
tur	Turkish	IMST	57	Turkic	
uig	Uyghur	UDT	40	Turkic	
ukr	Ukrainian	IU	116	Indo-European	
urd	Urdu	UDTB	138	Indo-European	
vie	Vietnamese	VTB	43	Austroasiatic	few features
\end{filecontents*}
\pgfplotstabletypeset[
  font=\tiny,
  column type=l,
  col sep = tab,
  string replace*={_}{.},
  every head row/.style={after row=\midrule},
  display columns/0/.style={string type},
  display columns/1/.style={string type},
  display columns/2/.style={string type},
  display columns/3/.style={numeric type,column type=r},
  display columns/4/.style={string type},
  display columns/5/.style={string type},
]
{treebanks.txt}
}

\begin{figure}
  \tikzsetnextfilename{fig-pca-rank}%
  \input{pca-rank-fig.tex}
  \vspace{-5mm}
  \caption{Visualization of all principal components.
    Languages are sorted according to the measure value in each panel
    (higher points indicate higher complexity).
    The original scale of each measure is given under each panel.
    The vertical gray lines represent the mean value of the measure.
  }\label{fig:pca-measures}
\end{figure}

\begin{table}
  \centering
  \caption{WALS features used.
    The column `Coverage' indicates the number of languages in our sample
    for which the feature is defined in WALS.}\label{tbl:wals}
\begin{tabular}{llr}
Feature ID& Description & Coverage\\
\midrule
22A & Inflectional Synthesis of the Verb & 18\\
26A & Prefixing vs. Suffixing in Inflectional Morphology & 34\\
27A & Reduplication & 22\\
28A & Case Syncretism & 19\\
29A & Syncretism in Verbal Person/Number Marking & 19\\
30A & Number of Genders & 18\\
33A & Coding of Nominal Plurality & 35\\
34A & Occurrence of Nominal Plurality & 23\\
37A & Definite Articles & 31\\
38A & Indefinite Articles & 29\\
49A & Number of Cases & 29\\
51A & Position of Case Affixes & 35\\
57A & Position of Pronominal Possessive Affixes & 27\\
59A & Possessive Classification & 18\\
65A & Perfective/Imperfective Aspect & 25\\
66A & The Past Tense & 25\\
67A & The Future Tense & 25\\
69A & Position of Tense-Aspect Affixes & 34\\
70A & The Morphological Imperative & 33\\
73A & The Optative & 25\\
74A & Situational Possibility & 28\\
75A & Epistemic Possibility & 28\\
78A & Coding of Evidentiality & 25\\
94A & Order of Adverbial Subordinator and Clause & 30\\
101A & Expression of Pronominal Subjects & 32\\
102A & Verbal Person Marking & 22\\
111A & Nonperiphrastic Causative Constructions & 19\\
112A & Negative Morphemes & 35\\
\end{tabular}
\end{table}

\end{document}